\documentclass{bmcart}

\usepackage[utf8]{inputenc} 

\def\includegraphics{}
\usepackage{graphicx}
\usepackage{tabularx}
\newcolumntype{Y}{>{\centering\arraybackslash}X}
\startlocaldefs
\endlocaldefs
\startlocaldefs
\endlocaldefs
\usepackage[usestackEOL]{stackengine}

\begin{document}

\begin{frontmatter}

\begin{fmbox}
\dochead{Research}


\title{Molecular generative model based on conditional variational autoencoder for de novo molecular design }


\author[
   addressref={aff1},                   
   email={ljchang94@kaist.ac.kr}   
]{\inits{}\fnm{Jaechang} \snm{Lim}}
\author[
   addressref={aff1},                   
   email={god_seongok@kaist.ac.kr}   
]{\inits{}\fnm{Seongok} \snm{Ryu}}
\author[
   addressref={aff1},                  
   email={starhouse@kaist.ac.kr}   
]{\inits{}\fnm{Jin Woo} \snm{Kim}}
\author[
   corref={aff1},
   addressref={aff1,aff2},                
   email={wooyoun@kaist.ac.kr}
]{\inits{}\fnm{Woo Youn} \snm{Kim}}


\address[id=aff1]{
  \orgname{Department of Chemistry, KAIST}, 
  \street{291 Daehak-ro},                     %
  \postcode{34141}                                
  \city{Daejeon},                              
  \cny{Republic of Korea}                                    
}
\address[id=aff2]{%
  \orgname{KI for Artificial Intelligence, KAIST},
  \street{291 Daehak-ro},
  \postcode{34141}
  \city{Daejeon},
  \cny{Republic of Korea}
}


\end{fmbox}


\begin{abstractbox}

\begin{abstract} 

We propose a molecular generative model based on the conditional variational autoencoder for de novo molecular design. It is specialized to control multiple molecular properties simultaneously by imposing them on a latent space. As a proof of concept, we demonstrate that it can be used to generate drug-like molecules with five target properties. We were also able to adjust a single property without changing the others and to manipulate it beyond the range of the dataset.
\end{abstract}




\end{abstractbox}
%

\end{frontmatter}



\section*{Introduction}
The ultimate goal of molecular design for new materials and drugs is to directly generate molecules with the desired properties. This is apparently challenging work because a molecular space is extraordinarily vast, discrete, and disorganized with diverse types of molecules. For instance, $10^{8}$ molecules have been synthesized,\cite{Kim2016} whereas it is estimated that there are $10^{23}$ to $10^{60}$ drug-like molecules.\cite{Polishchuk2013} Despite advances in experimental techniques, it is too demanding to find molecules suitable for specific applications only through experiments.

	Computer-aided molecular design has attracted much attention as a promising solution to overcome the experimental limitation.\cite{Shoichet2004,Scior2012,Cheng2012,Reymond2010} Fast calculation methods along with reasonable accuracy and very low cost enable high-throughput virtual screening to find molecules with target properties. A common strategy is to select computationally top molecules out of millions of molecules in a virtual library and then verify them experimentally, leading to a significant reduction in time and efforts. Molecules in the library may not meet the given criteria. In this case, traditional optimization methods such as a genetic algorithm can be used to further improve molecular properties beyond the criteria by structural modifications.\cite{Miyao2014,Hartenfeller2011,Rupakheti2015} However, they have a fundamental limitation in terms of efficiency because many trials and errors are inevitable to optimize molecular properties in a huge molecular space. 
    
	Recently emerging generative models based on deep learning techniques may offer a viable solution for more efficient molecular design. Gómez-Bombarelli et al. adopted a variational autoencoder\cite{Kingma2013} to optimize the molecular properties in a latent space in which molecules are expressed as a real vector.\cite{Gomez-Bombarelli2016} The key advantage of this method is that a gradient-based optimization becomes feasible because the latent space is continuous and differentiable. It has been successfully applied to improving the partition coefficient of drug candidates and the delayed fluorescent emission rate of organic light emitting diode candidates. Blaschke et al. employed the adversarial autoencoder\cite{Makhzani2015} (AAE) and the Bayesian optimization to generate ligands specific to the dopamine type 2 receptor.\cite{Blaschke2017} Kadurin et al. compared the VAE and AAE as a molecular generation model in terms of the reconstruction error and variability of the output molecular fingerprints.\cite{Kadurin2017} In addition to those autoencoder-based models, a generative model developed for natural language processing has also been used for molecular design.\cite{Bjerrum2017,Yuan2017,Segler2017,Gupta2017} Molecular structures can be expressed with SMILES. Then, this model learns the probability distribution of the next character of a given piece of SMILES. Yuan et al. designed potential inhibitors for a target protein and tested them in experiments.\cite{Yuan2017} Based on the natural language processing model, Segler et al. and Gupta et al. applied transfer learning to molecular design for certain biological activities.\cite{Segler2017,Gupta2017} This approach is especially useful when there is not enough data to train artificial neural networks in the normal way. Olivecrona et al., Guimaraes et al., and Jaques et al. proposed a reinforcement learning method to modify a pre-trained molecular generative model so as to impose several properties in molecules generated from the generative model.\cite{Olivecrona2017,Guimaraes2017,Jaques2016}
    
	We note that various molecular properties are correlated with each other. Therefore, adjusting one target property by structural modifications may cause an undesired change in other properties. To avoid this problem in rational molecular design, one has to control several properties at the same time. Here, we propose a molecular generative model using the conditional variational autoencoder (CVAE)\cite{Kingma2014} suitable for multivariable control. In addition to the advantages of using the latent space, our method can incorporate the information of molecular properties in the encoding process and manipulate them in the decoding process. 
    
As a proof of concept, we used the CVAE to generate drug-like molecules satisfying five target properties at the same time: molecular weight (MW), partition coefficient (LogP), number of hydrogen bond donor (HBD), number of hydrogen acceptor (HBA), and topological polar surface area (TPSA). We were able to produce a number of molecules with the specific values of the five target properties within a given range. It was also possible to adjust a single target without changing the others. Furthermore, we were able to generate molecules with properties beyond the range of the database. 
    
\section*{Method}

\subsection*{Conditional variational autoencoder (CVAE)}

We selected the CVAE as a molecular generator. It is one of the most popular generative models which generates objects similar to but not identical to a given dataset. In particular, it is distinguished from the VAE in that it can impose certain conditions in the encoding and decoding processes. To elucidate the difference between the two models, we compared their objective functions with one another. The objective function of the VAE is given by
\[
 E[\textrm{log}P(X|z)]-D_{KL}[Q(z|X)\parallel P(z)],
\]
where $E$ denotes an expectation value, $P$ and $Q$ are probability distributions, $D_{KL}$ is the Kullback-Leibler divergence, and $X$ and $z$ indicate the data and latent spaces, respectively. The first and second terms are also called the reconstruction error and the KL term, respectively. In an autoencoder, $Q(z|X)$ and $P(X|z)$ are approximated by an encoder and a decoder, respectively. A key difference of the CVAE from the VAE is to embed the conditional information in the objective function of the VAE, leading to the revised objective function as follow:
\[
 E[\textrm{log}P(X|z,c)]-D_{KL}[Q(z|X,c)\parallel P(z|c)],
\]
where $c$ denotes a condition vector. The condition vector $c$  is directly involved in the encoding and decoding processes. In our model, the molecular properties we want to control were represented as the condition vector. As a result, the CVAE can generate molecules with the target properties imposed by the condition vector.

Incorporating molecular properties in the VAE to generate molecules with desirable properties are also possible through a two-step model proposed by Gómez-Bombarelli et al. The VAE is trained jointly with an additional neural network for property prediction. Subsequently, a Gaussian process model creates a mapping from the resulting latent space to the associated molecular properties Finally, property optimization in the resulting latent space is performed by a gradient descent optimization method.

The key difference of our CVAE model from the jointly trained VAE model is that the molecular properties are directly incorporated into both the encoder and decoder. The resulting latent vector is composed of two parts: the first part is for the target molecular properties, while the second part involves the molecular structures and the other properties. Therefore, the desired molecular properties can be embedded in a target molecular structure simply by setting a condition vector. In other words, one can control the structure and the  properties independently except for some cases in which the properties are strongly coupled to a molecular scaffold. This is particularly useful to incorporate a certain property in a given molecule just with a marginal structure modification. After all, the CVAE is less sensitive to the continuity and smoothness of the latent space, because it does not require the derivative of the latent space with respect to the latent vector of the molecular structure. Another technical difference of the CVAE from the jointly trained VAE is that it does not need any further optimization process, which is inevitable in the jointly trained VAE for each different property value.

\subsection*{Molecular representation and model construction}
We represented molecules with SMILES codes to take advantage of state-of-the-art deep learning techniques that are specialized in dealing with texts and sequences. Each SMILES code was canonicalized for a unique molecular representation. One 'E' was padded on the end of the SMILES code to indicate the end of the string. Subsequently, each character including 'E' is represented with a one-hot vector, resulting in an input matrix. Each one-hot vector of the input matrix is transformed to an embedding vector with the size of 300, and then the input matrix is concatenated with a predefined condition vector. The first, second, and last entries of the condition vector are filled with information consisting of the MW, LogP, and TPSA, respectively, while the remaining two entries are labeled by the HBD and HBA as shown in Figure 1. The values of MW, logP, and TPSA are normalized from -1.0 to 1.0. HBD and HBA are expressed with a one-hot vector, because they are integer numbers.

The resulting matrix is subjected to the encoder of the CVAE to generate a latent vector. We adopted the so-called recurrent neural network (RNN) with an LSTM cell for both the encoder and decoder of the CVAE.\cite{Hochreiter1997} They were made of a 3-layer RNN with 500 hidden nodes on each layer. A softmax layer was used in each output of the decoder cell, and a cross entropy was used as the cost function of the reconstruction error. The latent vector concatenated with the condition vector becomes an input of the decoder at each time step of the RNN cell. Finally, the output vector of each decoder cell is transformed to a vector whose size is equal to that of the one-hot vector of the input matrix. The softmax activation function is applied to each transformed vector. The encoder and decoder are optimized to minimize the cost function of the CVAE. To generate a molecule with the target properties imposed by the condition vector, the cell of the RNN decoder was unrolled for 120 times. All characters before ‘E’ were taken in the stochastic write-out process, and if ‘E’ did not appear in the 120 characters, the result was considered as invalid. Each output vector of the decoder cell represents the probability distribution of the SMILES code characters and 'E'. Finally, the output vector is converted to a SMILES code. It should be noted that even a single wrong character in the resulting SMILES code gives rise to an invalid molecule. To increase the rate of valid SMILES codes, we used the stochastic write-out method which samples each character of SMILES according to a probability distribution. As a result, a single set of latent and condition vectors may give a number of different molecules. We performed 100 times  the stochastic write-out per one latent vector and took all valid molecules except duplicated ones for later analysis. 

\subsection*{Dataset and hyperparameters}
RDKit\cite{rdkit}, an open source cheminformatics package, was used for checking out the validity of the generated SMILES codes and calculating the five target properties of the molecules.

The total dataset is made of molecules randomly selected from the ZINC dataset.\cite{Irwin2012} Generally, with more data, the performance becomes better. Typical deep learning models need hundreds of thousands of data points. We checked out the convergence of the results with respect to the size of the data in our case. The use of 5,000,000 ZINC molecules did not increase both the validation and the success rates of generating molecules with the target properties compared to those from 500,000 ZINC molecules. Thus, we adopted the dataset of the 500,000 molecules,  80 \% of which were used for training, and the rest was used for the test. The distribution of the five target properties in the total dataset is shown in Figure 2. The learning rate was set to 0.0001 and exponentially decayed at a rate of 0.97. The model was trained until converged. In the performance evaluation of the CVAE, if each target property of the generated molecules was different from the given target value with the 10 \% error range of the average value of the total dataset, we regarded those molecules as successful.
 
 \begin{figure}[h!]
   \includegraphics[width=0.5\linewidth]{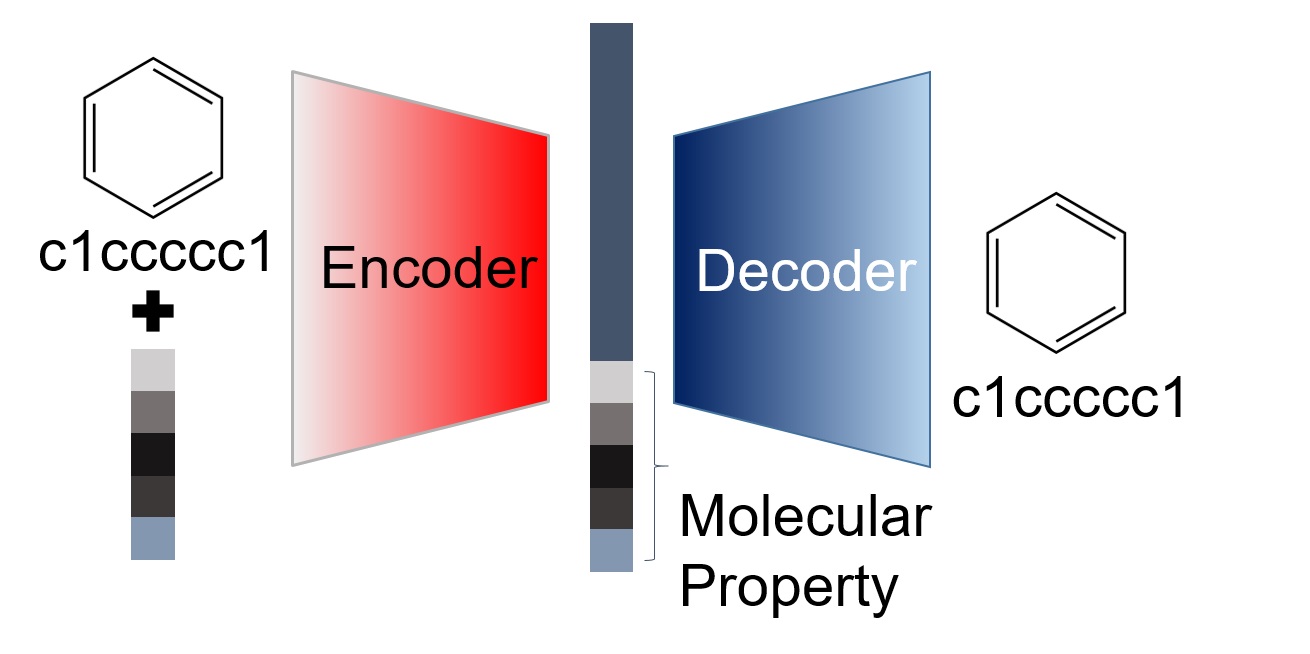}
   \caption{Schematic representation of conditional variational autoencoder for molecular design}
 \end{figure}
 
\begin{figure}[h!]
\includegraphics[width=0.5\linewidth]{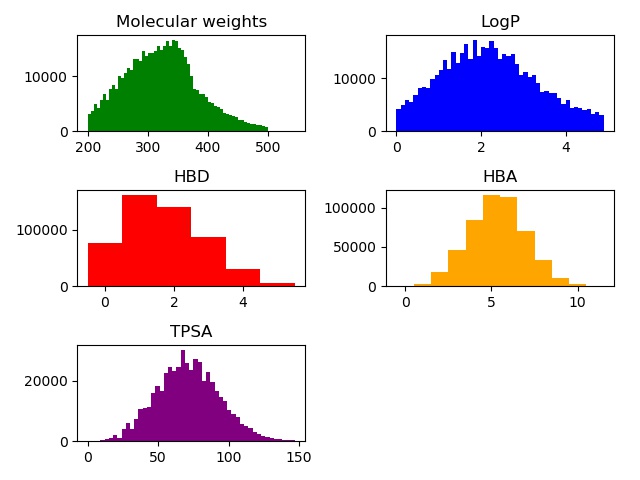}
   \caption{Distribution of molecular weight, LogP, HBD, HBA, and TPSA in the total dataset (500,000).}
 \end{figure}

\section*{Result}

 As the first application, we demonstrated that the CVAE method can generate molecules with specific values for the five target properties by applying it to Aspirin and Tamiflu. The values of the (MW, LogP, HBD, HBA, and TPSA) for Aspirin and Tamiflu are (180.04, 1.31, 1, 3, and 63.6) and (312.2, 1.285, 2, 5, and 90.64), respectively. The condition vector of each molecule was made by those values. Latent vectors to be concatenated with the condition vector were sampled by adding a Gaussian type noise to the latent vector of a molecule selected randomly in the training set. Figures 3A and 3B show nine molecules produced with the condition vector of Aspirin and Tamiflu, respectively. All of them had similar properties to those of Aspirin and Tamiflu within an error range of 10\%, respectively. However, the molecular structures in Figure 3 are considerably different from those of the original molecules because of the latent vectors chosen randomly from the training set.

The second application was to generate molecules similar in both properties and structure to the mother molecule by sampling latent vectors around that of the mother. Figure 4 shows the molecules generated in such a way from Aspirin. They look very similar to Aspirin and also have similar properties with those of Aspirin within an error range of 10 \%.

 \begin{figure}[h!]
 \includegraphics[width=0.5\linewidth]{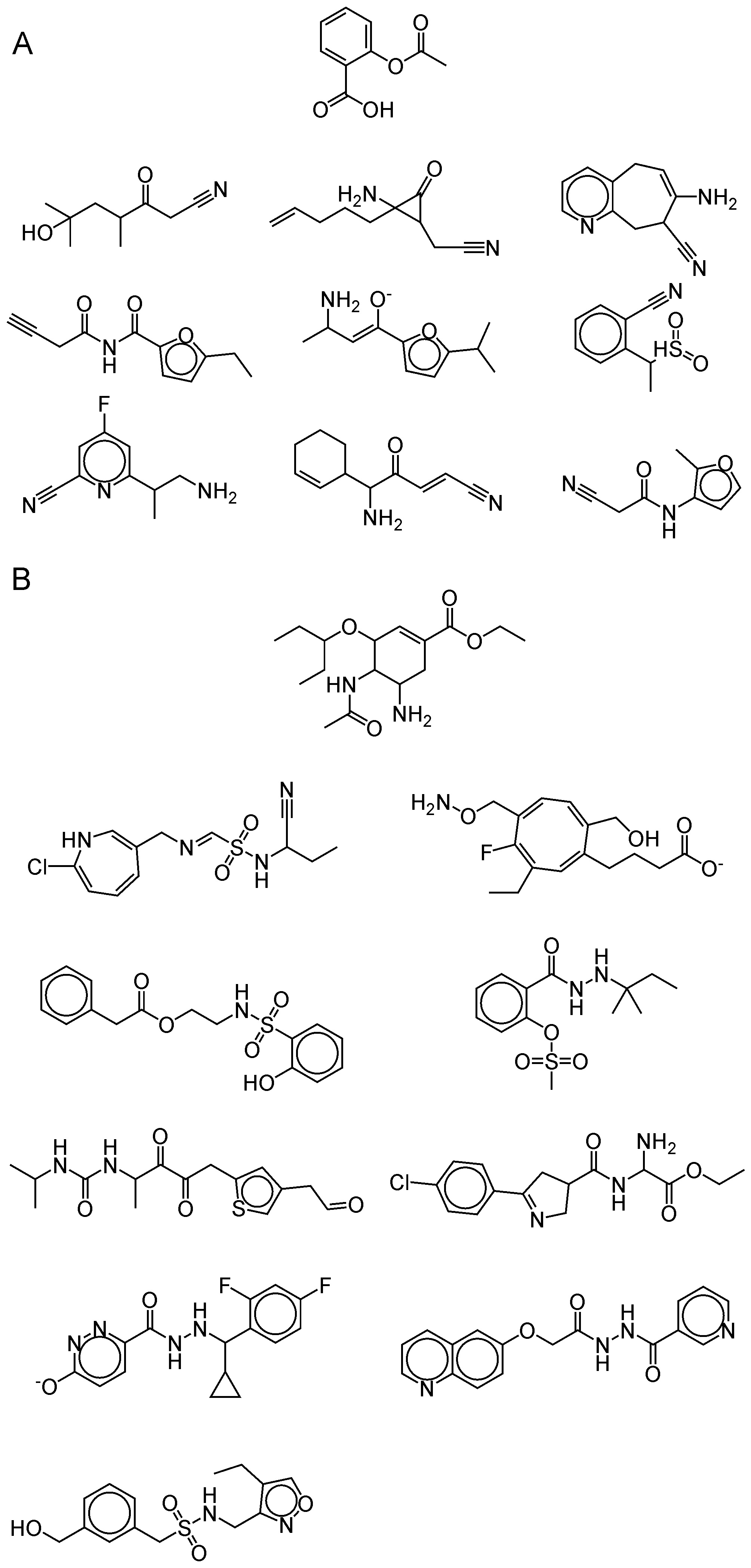}
   \caption{Molecules generated by the CVAE with the condition vector made of the five target properties of (A) Aspirin and (B) Tamiflu. }
 \end{figure}
 
  \begin{figure}[h!]
 \includegraphics[width=0.5\linewidth]{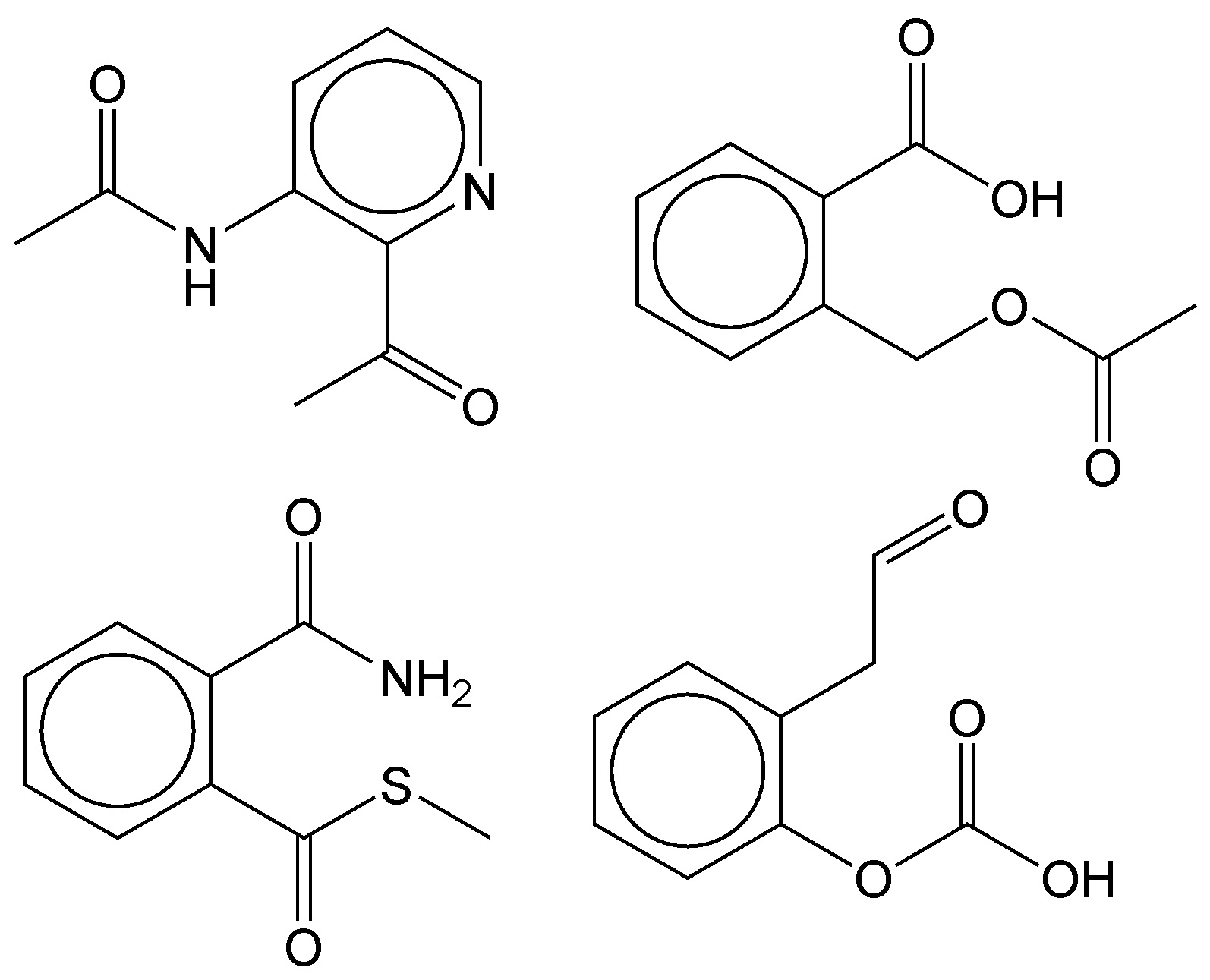}
   \caption{Molecules generated by the CVAE with the condition vector made of the five target properties of Aspirin and the latent vector slightly modified from that of Aspirin. }
 \end{figure}
 
As the third case study, we tested whether the CVAE method can change only a single property without changing the others. The condition vector was constructed with the MW, HBD, HBA, and TPSA of Tamiflu, and we varied LogP from 0.0 to 3.0. Latent vectors were sampled around that of Tamiflu. Figure 5 shows the result. All the molecules have similar properties to the original ones except LogP as desired. The molecules from the top left to the bottom right have gradually increasing LogP values from -0.23 to 3.55. In some cases, however, such a delicate control of individual properties was not possible. For instance, we could not generate molecules with a LogP beyond 4.0. It is probably because LogP is not completly independent from the other four properties, so a substantial change in LogP entails a change in the others. Moreover, it was difficult to adjust the MW and TPSA independently because the MW and TPSA are highly correlated with one another.
\begin{figure}[b!]
 \includegraphics[width=0.5\linewidth]{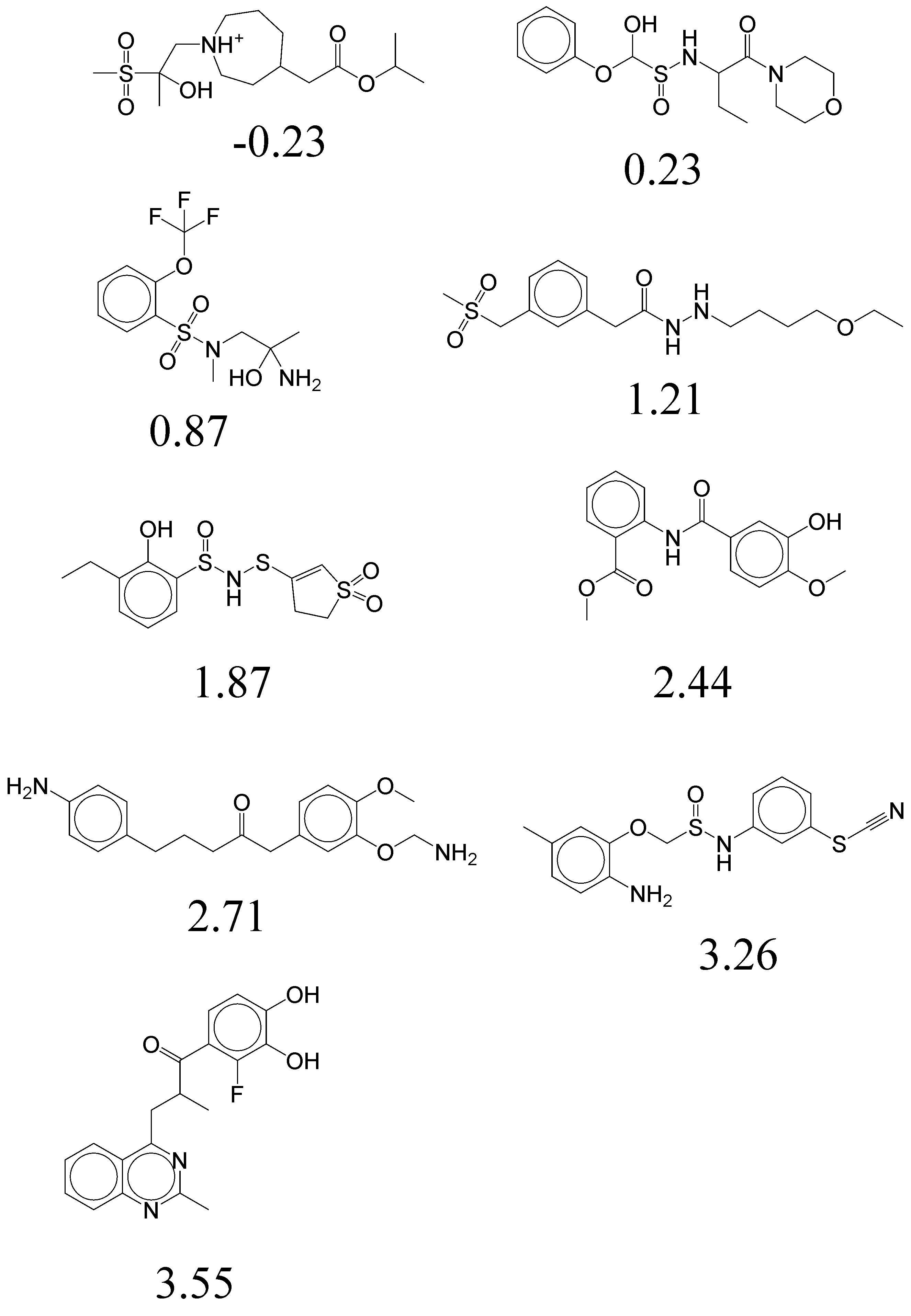}
   \caption{Molecules generated by the CVAE with the condition vector made of MW, HBD, HBA, and TPSA of Tamiflu and continuously changing LogP.}
 \end{figure}

Finally, we investigated the possibility to change a specific molecular property beyond the range of a training set. Latent vectors were sampled around molecules in the training set. In the condition vector, the five properties were given randomly except for a single target property. The target property was set to 10 \% larger than its maximum value in the training set (e.g., 5.5 for LogP and 165.5 for TPSA). Figure 6 shows the resulting molecules. Indeed, it was able to generate molecules with a LogP larger than 5.5 (Figure 6A) and molecules with a TPSA larger than 165.5 (Figure 6B). We compared the distribution of the LogP and TPSA for 1,000 randomly selected molecules from the training set and 1,000 generated molecules with property values outside of the range of the dataset (toward larger values). Figure7 shows that the distribution of the target properties are shifted to larger values, leading to an increased ratio of molecules with property values outside of the range. The rate of valid molecules is relatively low compared to the case of generating molecules with property values in the range of the dataset.
\begin{figure}[h!]
 \includegraphics[width=0.5\linewidth]{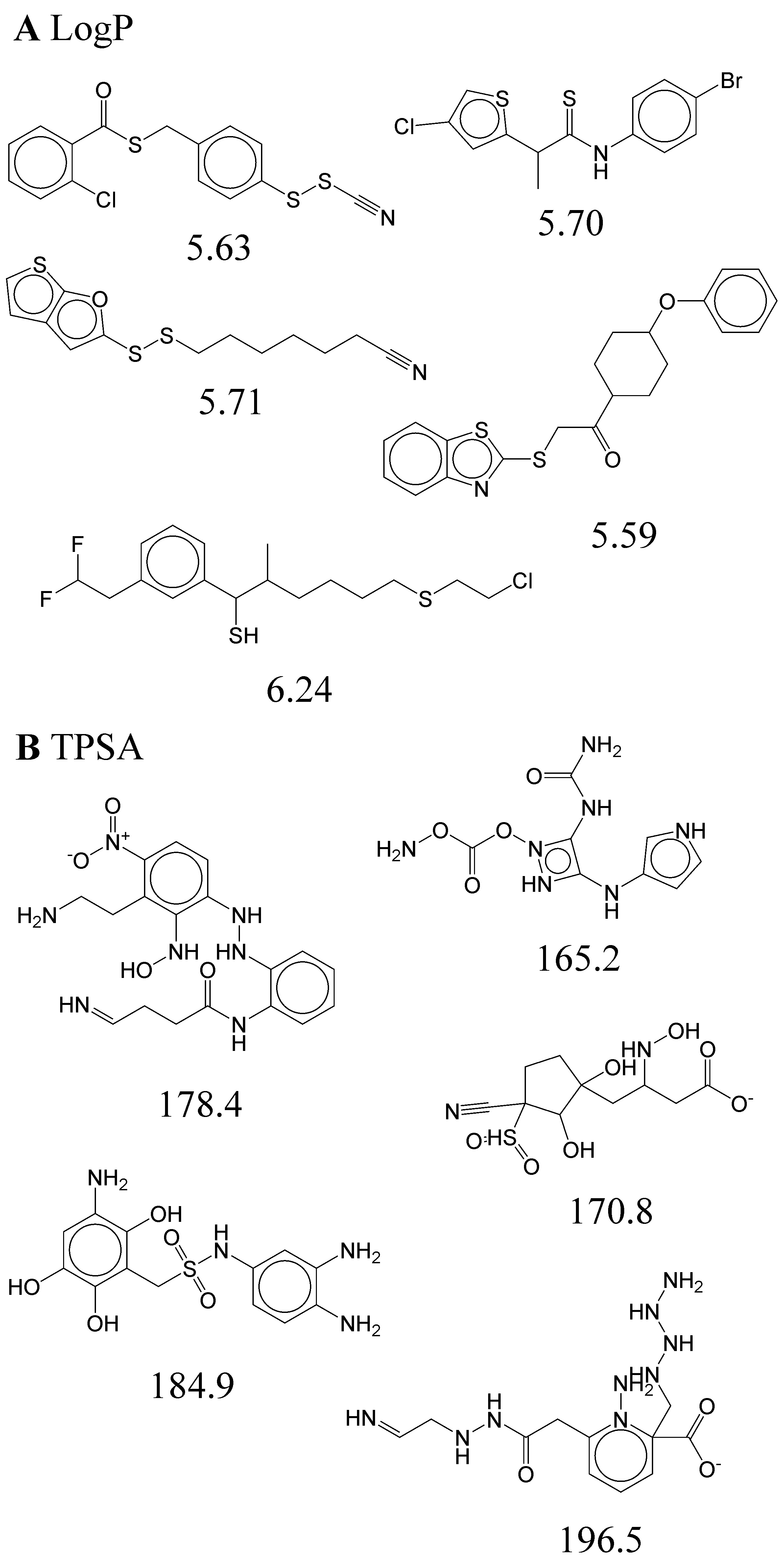}
   \caption{(A) Molecules with LogP larger than 5.5. (B) Molecules with TPSA larger than 165.}
 \end{figure}
 
 \begin{figure}[h!]
 \includegraphics[width=0.5\linewidth]{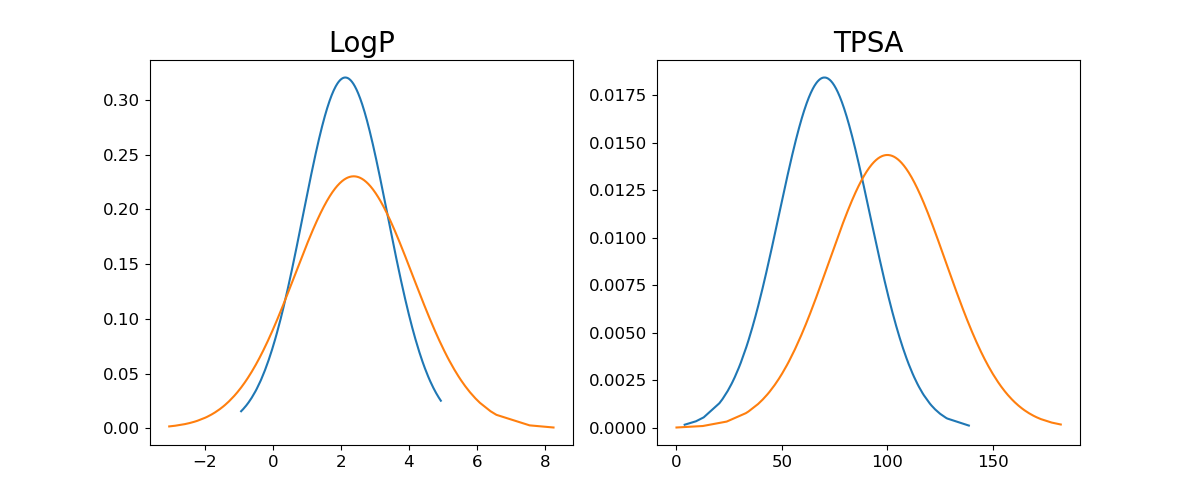}
   \caption{Distribution of A) LogP and B) TPSA for 1000 randomly selected molecules in training set and 1000 generated molecules with LogP and TPSA outside of the range of the dataset, respectively.}
 \end{figure}
 
\begin{table}[h!]
\caption{Numbers of attempts and valid molecules for generating 100 molecules whose five properties are the same with those of Aspirin, Tamiflu, Lenalidomide, Rivaroxaban, and Pregabalin.}
      \begin{tabular}{ccccccc}
        \hline
 {} & \multicolumn{3}{c}{\Centerstack{condition}}   & \multicolumn{3}{c}{\Centerstack{random}}\\ \hline
 		{} & \Centerstack{attempts} &\Centerstack{number of\\valid \\molecules} &\Centerstack{success rate \\(100/attempts, \\ \%)} &\Centerstack{attempts} & \Centerstack{number of \\ valid \\molecules} &\Centerstack{success rate \\(100/attempts, \\ \%)}\\ \hline
        Aspirin & 28,840 & 32,567 & 0.34 & 758,199 & 711,660 & 0.014\\
        Tamiflu & 15,960 & 34,696 & 0.62 & 798,183 & 741,960 & 0.013\\
        Lenalidomide & 502,00 & 89,230 & 0.19 & 865,695 & 822,060 & 0.012\\ 
        Rivaroxaban & 92,620 & 47,574 &  0.11& 866,205 & 817,800 & 0.012\\ 
        Pregabalin & 77,680 & 84,371 & 0.13 & 782,010 & 723,360 & 0.014\\ \hline
      \end{tabular}
\end{table}

We analyzed the latent space constructed by the CVAE. Two principle axes were extracted by principal component analysis. Figure 8 shows the two components of the latent vectors of 1000 randomly selected molecules from the test set with their LogP and TPSA values. Molecules with similar properties are likely located around a same region of the latent space in the jointly-trained VAE. In our CVAE model, the latent vector is comprised of two parts as explained in the method section. Therefore, a specific region in the latent space does not necessarily have a correlation with the target molecular properties which are controlled by the condition vector. This is good because the separation of information enables a more flexible control of the molecular structure and properties when generating new molecules.
\begin{figure}[h!]
 \includegraphics[width=0.5\linewidth]{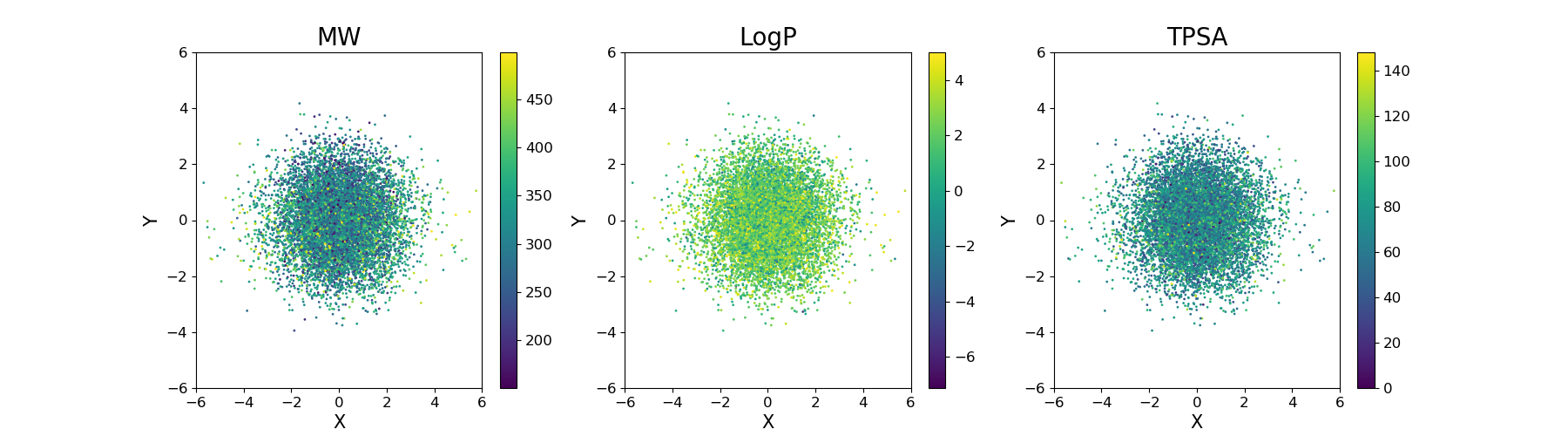}
   \caption{The latent space of 1000 randomly selected molecules with MW, LogP and TPSA values.}
 \end{figure}

Apart from the successful applications of the CVAE method, it has a drawback that should be resolved. The success rate of generating desirable molecules is very low. We tested how many attempts are required to generate 100 molecules with the five desired properties and how many valid molecules are generated during those attempts. We also compared when the condition vector is set randomly or to target properties to show the effect of the condition vector for generating desirable molecules.

Table 1 summarizes the number of attempts for generating 100 molecules whose five properties are same as those of aspirin, Tamiflu, Lenalidomide, Rivaroxaban, and Pregabalin, respectively. Lenalidomide, Rivaroxaban, and Pregabalin are top selling small molecule drugs in 2016.\cite{drug} In Table 1, 'condition' means that the condition vector was set as the five properties of the target molecules, whereas 'random' means that the condition vector was randomly made. The number of valid molecules in Table 1 indicates the number of valid molecules generated during the attempts to create molecules with the five desired properties. For example, 100 aspirin-like molecules and 32,567 valid moleculces were obtained from 28,840 attempts to create aspirin-like molecules. The reason why the number of valid molecules is larger than the number of attempts is that the stochastic write-out process is performed 100 times for each attempt. All successful molecules (100 per each target molecule) are reported in the Supporting Information. It should be noted that the success rate dramatically dropped when the condition vector is randomly set. It clearly manifests that the successful molecules generated by the CVAE in the example studies were not the result of many random trials.

\begin{table}[h!]
\caption{Number of generation attempts and number of valid molecules for three different sampling methods of latent vectors. The generation process was continued until 100 molecules with the five target properties were successfully created from a single target molecule, and it was repeated for 100 target molecules selected randomly from the ZINC dataset. The table shows the average values over the 100 target molecules.}
      \begin{tabular}{cccc}
        \hline
 sampling method & \Centerstack{average number of \\ valid molecules}    & \Centerstack{average number of \\ attempts} &\Centerstack{success rate \\(100/attempts, \%)}
 \\ \hline
 	        around target molecules & 67640.5 & 199340.1 & 0.05\\
        around known molecules &31799.4 & 21659.9 & 0.46 \\
        random &  50316.4 & 78888.2 & 0.12\\ \hline
       
      \end{tabular}
\end{table}

We further analyzed the performance of the CVAE by investigating the change in the success rate and the number of valid molecules according to latent vector sampling methods. We employed three different sampling methods: random, around the latent vectors of known molecules, and around the latent vectors of target molecules. For all the sampling methods, the condition vector was constructed using the five properties of the target molecules. The generation process was continued until 100 molecules with the five target properties were successfully created from a single target molecule, and it was repeated for 100 target molecules selected randomly from the ZINC dataset. Table 2 shows the average values for the success rate and the number of valid molecules over the 100 target molecules. 
It was unexpected that sampling latent vectors around a target molecule was the most ineffective in terms of the success rate and valid molecules because of the high rate of duplicated molecules. In this case, the structure of the generated molecules was very similar to that of the target molecule shown in Figure 4. Sampling latent vectors around those of known molecules performed best. Because the known molecules were randomly selected from the ZINC set, their structures and properties would be considerably different from those of a target molecule. Nonetheless, we were able to generate molecules with the desired properties from those latent vectors with a relatively high success rate. It manifests that the condition vector appropriately modified the molecular structures to have the target properties. Finally, it was also possible to generate desirable molecules from completely random latent vectors but with a low success rate.

We suspect that at some part the overall low success rates regardless of the latent vector sampling methods are due to the strong correlation between the five target properties. In addition, it is known that the discrete nature of SMILES causes a high rate of invalid molecules in the decoding process from latent vectors to molecules.\cite{Kusner2017} The stochastic write-out method circumvents this problem, but more fundamental solutions should be devised. More severely, SMILES does not have the 3D conformational information of molecular structures. Therefore, it must have limitations in applications in which conformational effects are critical. Molecular graph representation incorporating conformational information can be a promising alternative. Encoding molecular graphs seems to be straightforward, but decoding from a latent space to molecular graphs is still an open problem. Recently, significant progress along this line has been made.\cite{Wang2017,You2018,Jin2018} Such a better molecular representation may also improve the success rate of molecular generation. We expect that the success rate may be further improved by using the grammar variational autoencoder\cite{Kusner2017} and the reinforcement learning.\cite{Olivecrona2017,Guimaraes2017}
 
 \section*{Conclusion}
We proposed a new molecular design strategy based on the conditional variational autoencoder. Instead of high-throughput virtual screening, our method as one of the deep learning-based generative models directly produces molecules with desirable target properties. In particular, its strength is controlling multiple target properties simultaneously by imposing them on a condition vector. We demonstrated that it was possible to generate drug-like molecules with specific values for the five target properties (MW, LogP, HBD, HBA, and TPSA) within an error range of 10\%. In addition, we were able to selectively control LogP without changing the other properties and to increase a specific property beyond the range of the training set. Thus, this new method has attractive applicability for efficient molecular design.


\begin{backmatter}

\section*{Competing interests}
  The authors declare that they have no competing interests.

\section*{Author's contributions}
    Jaechang Lim, Seongok Ryu, Jin Woo Kim, and Woo Youn Kim organized this work. Jaechang Lim and Woo Youn Kim wrote the paper 

\section*{Acknowledgements}
  This work was supported by Basic Science Research Programs through the National Research Foundation of Korea (NRF) funded by the Ministry of Science, ICT and Future Planning (NRF-2017R1E1A1A01078109).


\bibliographystyle{spphys.bst} 
\bibliography{reference.bib}







\end{backmatter}
\end{document}